\documentclass{article}

\usepackage[english]{babel}
\usepackage{graphicx}
\usepackage{amsmath,amssymb,stmaryrd,mathtools}
\usepackage[utf8]{inputenc} 
\usepackage{graphicx}
\usepackage{algorithmic,algorithm}
\usepackage{multirow}
\usepackage{pdflscape}
\usepackage{url}
\begin{document}

\title{Boolean Networks Design by Genetic Algorithms}

\author{Andrea Roli\thanks{Corresponding author. Email: \texttt{andrea.roli@unibo.it}}, Cristian Arcaroli, Marco Lazzarini, Stefano Benedettini\\DEIS--Cesena, \textit{Alma Mater Studiorum} Universit\`{a} di Bologna, Italy}

\maketitle

\begin{abstract}
We present and discuss the results of an experimental analysis in the design of Boolean networks by means of genetic algorithms. A population of networks is evolved with the aim of finding a network such that the attractor it reaches is of required length $l$. In general, any target can be defined, provided that it is possible to model the task as an optimisation problem over the space of networks.
We experiment with different initial conditions for the networks, namely in ordered, chaotic and critical regions, and also with different target length values. Results show that all kinds of initial networks can attain the desired goal, but with different success ratios: initial populations composed of critical or chaotic networks are more likely to reach the target. Moreover, the evolution starting from critical networks achieves the best overall performance. This study is the first step toward the use of search algorithms as tools for automatically design Boolean networks with required properties.
\end{abstract}

\section{Introduction}

The design of complex systems is one of the main challenges in scientific and engineering disciplines. Model synthesis, identification and tuning, reverse engineering of biological and social networks, design of self-organising artificial systems are just some of the areas in which scientists are asked to face this issue. Such systems and models are mostly designed and tuned by means of automatic procedures, some of which can be ascribed to the class of \textit{search methods}. A prominent example of these approaches are evolutionary computation techniques, for instance for designing robotic systems~\cite{nolfi-evorobot-book}.

In this paper we present and discuss results of a preliminary study in the context of automatic design of Boolean networks via Genetic algorithms. Boolean networks (BNs) have been introduced by Kauffman as a model for genetic regulatory networks~\cite{Kauf93} and have been also studied as computational learning systems~\cite{Kauf91-sciam,Dori94}. 
The first study on the evolution of BNs can be found in~\cite{kauff1986-adaptive}, in which the authors apply a simple evolutionary algorithm to evolve BNs with an attractor containing a target state. A follow-up of that seminal work is that of Lemke at al.~\cite{lemke2001}, in which the fitness function accounts also for a desired attractor length. These studies are mainly an investigation of how evolution performs over BNs and raise interesting and fundamental questions on the search landscape structure and the evolutionary dynamics depending on network structural characteristics. More recently, works addressing the evolvability of robustness in BNs have been presented~\cite{aldana-robust,szejka2008-evolution-of-canalizing,braunewell2008-reliability}. In the same direction is a recent paper,~\cite{esmaeili2008}, in which the global fitness function is defined as the sum of single functions, each related to a network parameter somehow linked to network robustness (e.g.,  number and length of attractors).

Despite the amount of analytical studies on the properties of BNs and their effectiveness in capturing fundamental genetic phenomena, little effort has been received so far concerning their synthesis. The availability of tools for automatic design of BNs would make it possible to design and tune BNs for applications in genetics, as genetic regulatory network models~\cite{kauff2004-ensemble}, and robotics, as multi-functional controllers.

This contribution is a first step toward the development of a family of tools for automatic design of BNs and discrete dynamical systems in general. We first introduce preliminary definitions and concepts in Sections~\ref{sec:bn} and~\ref{sec:ga}; experimental settings and results are described in Section~\ref{sec:results}. We then conclude with an outline of the agenda for future research in Section~\ref{sec:agenda}.

\section{Boolean networks}
\label{sec:bn}

BNs have been firstly introduced by Kauffman~\cite{Kauf93} and subsequently received considerable attention in the composite community of complex systems. Recent advances in this research field can be mainly found in works addressing themes in genetic regulatory networks or investigating properties of BNs themselves~\cite{aldana-robust,fretter-response,ribeiro-mutual-information,SerVilGraKau-theorbio}.

A BN is a discrete-state and discrete-time dynamical system defined by a directed graph of $N$ nodes, each associated to a Boolean variable $x_i$, $i = 1,\ldots,N$, and a Boolean function $f_i(x_{i_1},\ldots,x_{i_{K_i}}$), where $K_i$ is the number of inputs of node $i$. Often, $K_i$ is chosen to be equal to a constant value $K$ for every $i$. The arguments of the Boolean function $f_i$ are the nodes whose outgoing arcs are connected to node $i$. The state of the system at time $t$, $t \in \mathbb{N}$, is defined by the array of the $N$ Boolean variable values at time $t$: $s(t) = (x_1(t), \ldots, x_N(t))$. The most studied dynamics for BNs is \textit{synchronous}, i.e., nodes update their states in parallel, and \textit{deterministic}. However, many variants exists, including asynchronous and probabilistic update rules~\cite{Gershenson2004c}.

In this work, we consider networks ruled by synchronous and deterministic dynamics. Given this setting, the network trajectory in the $N$-dimensional state space is a sequence of states composing a \textit{transient}, possibly empty, followed by an \textit{attractor}, that is a cycle of length $l \in [1,\ldots, 2^N]$. States that belong to a trajectory ending at attractor $A_i$ are said to be members of the \textit{basin of attraction} of $A_i$. When BNs are employed as genetic regulatory network models, attractors assume a notable relevance as they can be interpreted as cellular types~\cite{huang-breast-desease2007}. 

A special category of BNs that has received particular attention is that of Random BNs, which can capture relevant phenomena in genetic and cellular mechanisms and complex systems in general. Random BNs (RBNs) are usually generated by choosing at random $K$ inputs per node and by defining the Boolean functions by assigning to each entry of the truth tables a 1 with probability $p$ and a 0 with probability $1-p$. Parameter $p$ is called \textit{homogeneity} or \textit{bias}. Depending on the values of $K$ and $p$ the dynamics of RBNs is \textit{ordered} or \textit{chaotic}. In the first case, the majority of nodes in the attractor is frozen and any moderate-size perturbation is rapidly dampened and the network returns to its original attractor. Conversely, in chaotic dynamics, attractor cycles are very long and the system is extremely sensitive to small perturbations: slightly different initial states lead to divergent trajectories in the state space. RBNs temporal evolution undergo a second order phase transition between order and chaos, governed by the following relation between $K$ and $p$: $K_c = [2 p_c (1 - p_c)]^{-1}$, where the subscript $c$ denotes the critical values~\cite{derrida1986}. Networks along the \textit{critical} line have important properties, such as the capability of achieving the best balance between evolvability and robustness~\cite{aldana-robust} and maximising the average mutual information among nodes~\cite{ribeiro-mutual-information}. Hence the conjecture that living cells, and living systems in general, are critical~\cite{aldana-macrophage}.

In this work we are interested in designing BNs such that the attractor they reach from a given initial state has a target length: this represents just one of numerous examples of requirements we may want a BN to satisfy. Nevertheless, since attractor length depends on the main properties of BNs, this goal enables us to address some of the most relevant issues in BN design.

\section{Genetic algorithms}
\label{sec:ga}

Genetic algorithms (GAs) belongs to the family of evolutionary computation methods and have been successfully applied as search techniques since several decades~\cite{Holland75,Goldberg89,Michalewicz96}. Inspired by Darwin's theory of selection and natural evolution, a GA evolves a population of candidate solutions to a problem by iteratively selecting, recombining and modifying them. The driving force of the algorithm is selection, that biases search toward the \textit{fittest} solutions, i.e., those with the highest objective function value. 
Algorithm~\ref{algo:ec-basic} shows the basic structure of a GA.

\begin{algorithm}
\caption{Genetic Algorithm}
\label{algo:ec-basic}
\begin{algorithmic}
\STATE $P \leftarrow$ {\sf GenerateInitialPopulation()}
\STATE {\sf Evaluate($P$)}
\WHILE{termination conditions not met}
  \STATE $P' \leftarrow$ {\sf Recombine($P$)}
  \STATE $P'' \leftarrow$ {\sf Mutate($P'$)}
  \STATE {\sf Evaluate($P''$)}
  \STATE $P \leftarrow$ {\sf Select($P'' \cup P$)}
\ENDWHILE
\end{algorithmic}
\end{algorithm}

The function {\sf Evaluate($P$)} computes the fitness of each individual of population $P$. The fitness function is positively correlated with the objective function, that quantifies the quality of a candidate solution and it is usually normalised in the range [0,1]. In the next section, we detail the specific genetic algorithm used in our experiments.

\section{Experimental analysis}
\label{sec:results}

The long term aim of this study is the definition and implementation of automatic procedures and methodologies for designing BNs and similar systems. The availability of such procedures would make it possible to perform inference of real genetic networks and to study the effects of evolution on simple genetic models~\cite{kauff2004-ensemble}. Furthermore, a promising yet uninvestigated research area consists of using BNs to control autonomous systems: the same BN-controller can produce different behaviours, depending on the attractor it is traversing. The actual behaviours have to be encoded into a proper sequence of states, hence the need for a procedure for defining the network according to the requirements.

In this work, our goal is to investigate the possibility of evolving BNs by GAs so as to obtain a network able to reach an attractor of a desired period with a trajectory starting from a given initial state $s_0$. The questions that we want to address are the following:

\begin{itemize}
  \item[\parbox{0.5cm}{(a)}] Is it possible to guide evolution in such a way to succeed in the goal? What is the probability of reaching the target? (i.e., how robust is the automatic design procedure?)
  \item[\parbox{0.5cm}{(b)}] Are there differences across network parameters? Are there networks that are \textit{easier} to evolve?
  \item[\parbox{0.5cm}{(c)}] Which are the most difficult or the easiest targets to be reached?
  \item[\parbox{0.5cm}{(d)}] What is the influence of GA parameters?
  \item[\parbox{0.5cm}{(e)}] What are the effect of the evolution on networks structure?
\end{itemize}

In the remainder of this section we detail the experimental settings and report and discuss the experimental results.

\subsection{Experimental settings}

Experiments are run with networks of 100 nodes and $K = 3$. The initial state is chosen at random and the target attractor lengths are 1, 10, 50, 100, 500, 800. Networks composing the initial population are constructed by randomly assigning inputs, without self-inputs; Boolean functions are defined by assigning truth values biased by homogeneity values equal to 0.85 (ordered), 0.788675 (critical) and 0.5 (chaotic), in three different experiment series, respectively. However, Boolean functions homogeneity of single individuals can change during evolution because the initial distribution of 1s and 0s can be changed by the genetic operators. For efficiency reasons, the temporal evolution of each network is simulated for at most 1000 steps: if an attractor is not reached in this limit, a fitness value of 0 is returned.
The individuals of the GA are encoded as a tuple of $N$ binary vectors of size $2^K$, each defining the Boolean function of a node. Thus, only Boolean functions of a network are evolved and the connections are kept constant. The recombination operator is a one-point crossover and mutation is a single-variable flip. Both operators are applied chromosome-wise. The fitness function is defined as $F(net) = (1 + |l - l_t|)^{-1}$, where $l$ is the length of the attractor the individual network reached and $l_t$ is the target length.
The remaining parameters of the GA have been chosen as reported in Table~\ref{tab:ga-params}, in which a summary of experimental parameter values is provided. All the possible combinations of the values reported have been tested.

\begin{table}[t]
\caption{Summary of experimental parameter values. All the possible combinations of the values reported have been tested.}
\begin{center}
\begin{scriptsize}
\begin{tabular}{|c|c|c|c|c|c|c|c|c|}
\hline
$N$ & $K$ & $p$ & attractor & population & number of   & mutation / crossover & number of\\
    &     &     & length    & size       & generations & rate        & runs\\
\hline
\hline
      &   &           & 1   &     &     &             & \\
      &   &           & 10  &     &     &   &\\
      &   & 0.5       & 50  &     &     & 0.5 / 0.9  & \\
100   & 3 & 0.788675  & 100 &  80 & 200 & 0.5 / 0.0   & 100\\
      &   & 0.85      & 500 &     &     &  0.1 / 0.9  & \\
      &   &           & 800 &     &     &           & \\
\hline
\end{tabular}
\end{scriptsize}
\end{center}
\label{tab:ga-params}
\end{table}

BNs have been simulated with a BN simulator developed by the group of Artificial Intelligence and Complex Systems of DEIS-Cesena, further extended into the BN Simulation Toolkit~\cite{BNToolkit} and the GA has been implemented with GAUL~\cite{GAUL}. All experiments have been performed on a 2.4 GHz Intel Core 2 Quad with 4MB of cache and 2GB of RAM, running with Linux Ubuntu 8.10.

\subsection{Performance comparison}

\begin{figure}
\begin{center}
 \includegraphics[scale=0.46,angle=0]{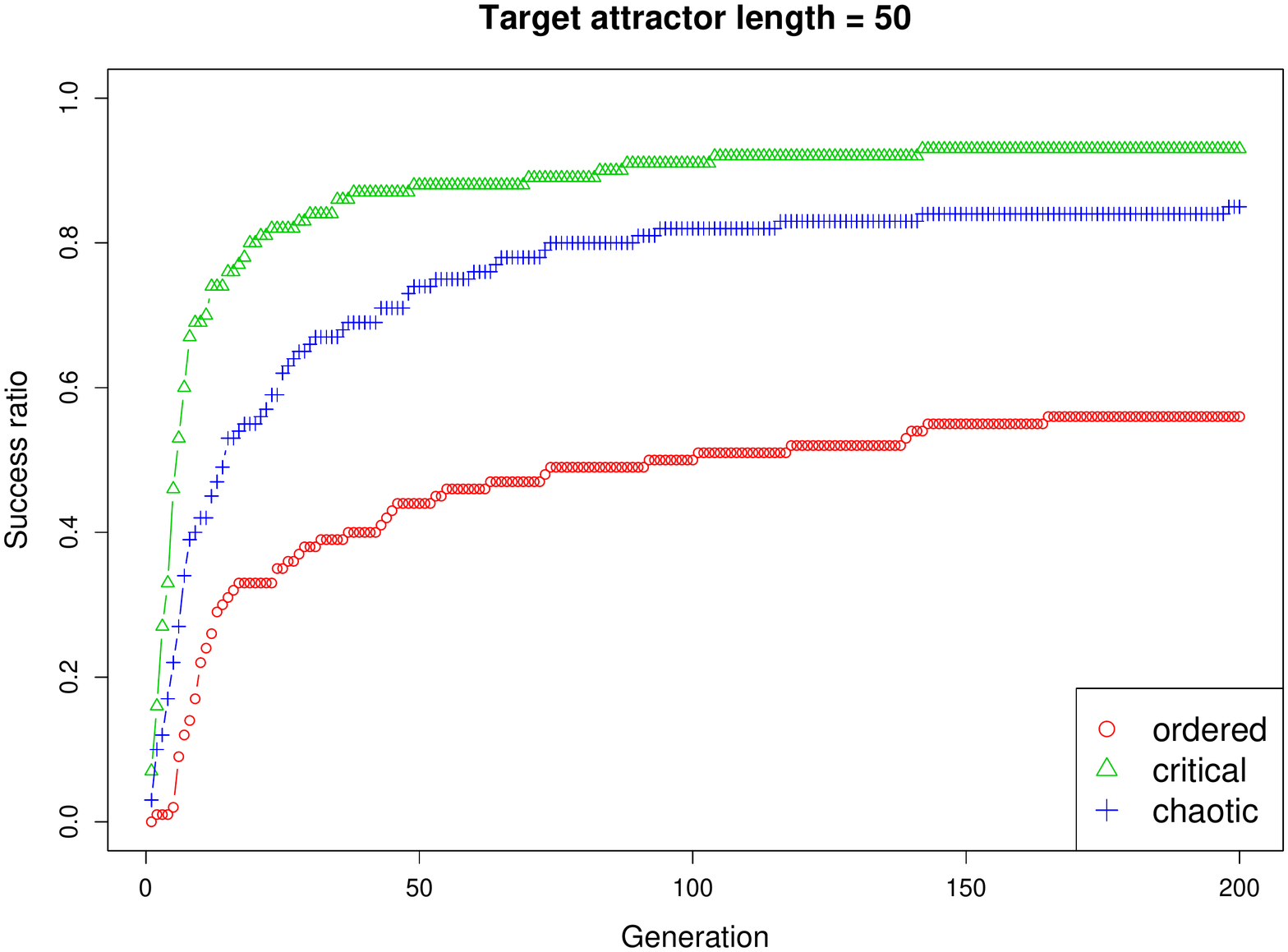}
 \includegraphics[scale=0.46,angle=0]{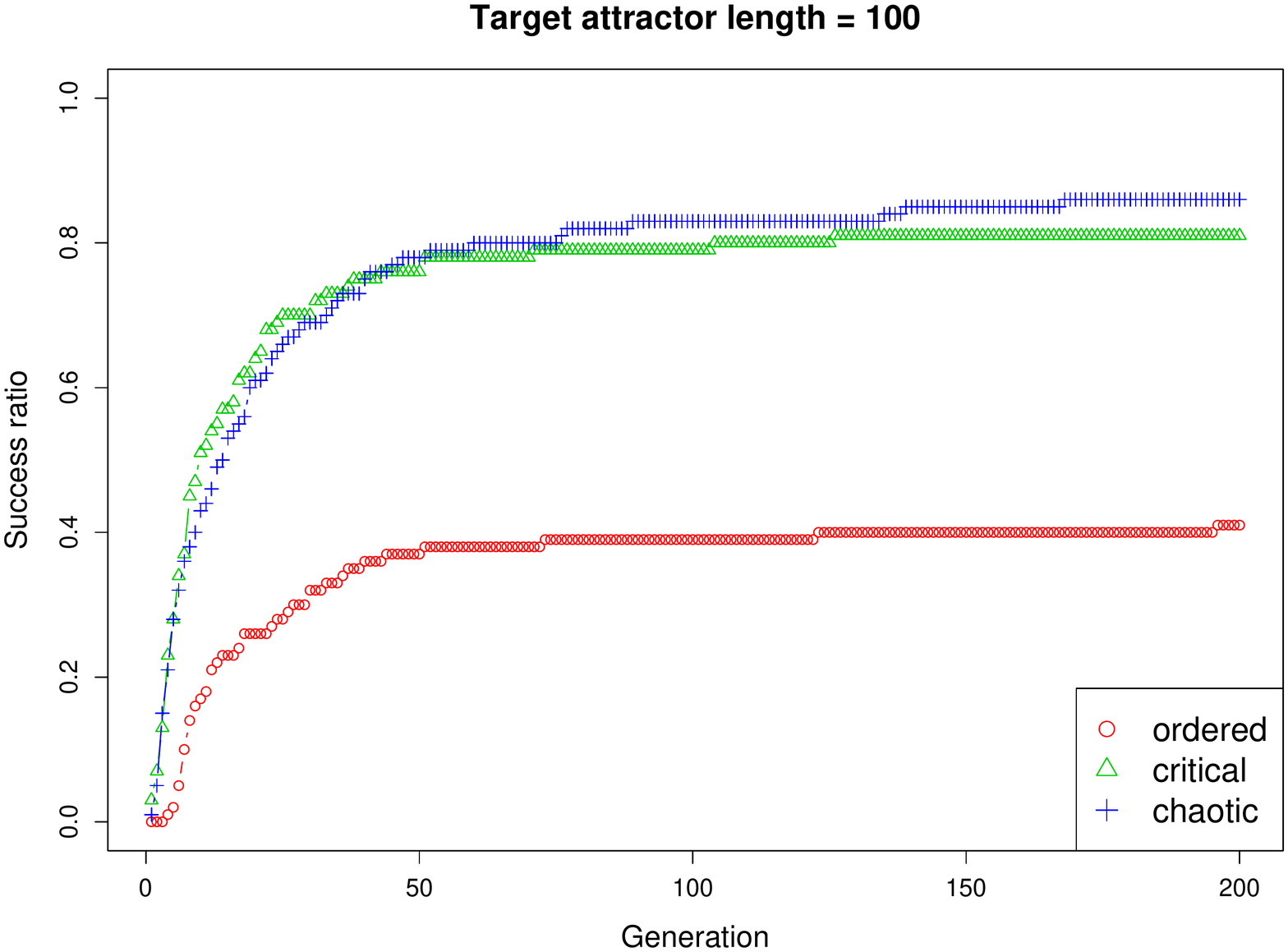}
\end{center}
\caption{Success ratio vs. generations. The comparison is made among the three initial network classes. Target attractor lengths equal to 50 and 100.}
\label{fig:plot-rld-a}
\end{figure}

\begin{figure}
\begin{center}
 \includegraphics[scale=0.46,angle=0]{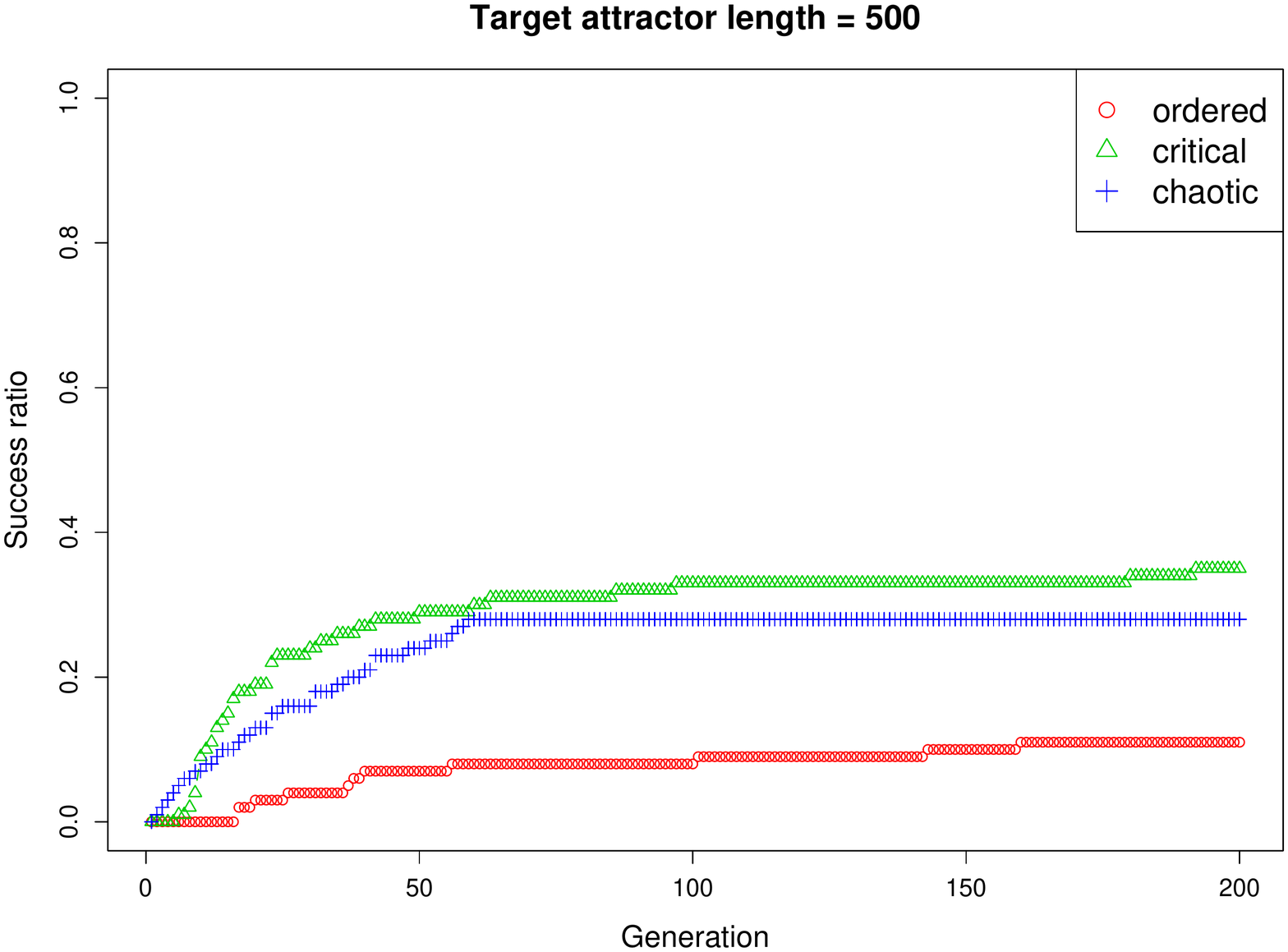}
 \includegraphics[scale=0.46,angle=0]{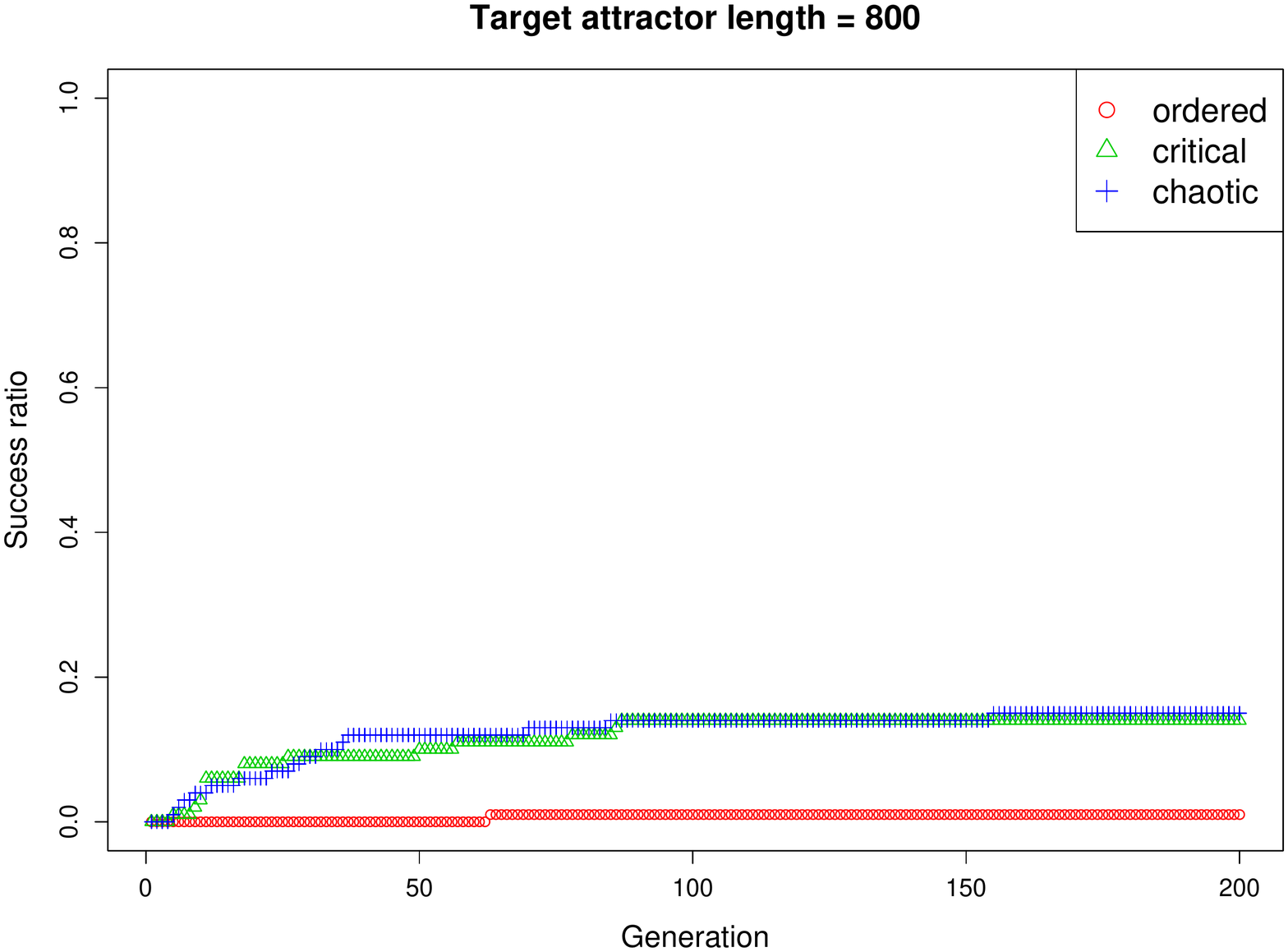}
\end{center}
\caption{Success ratio vs. generations. The comparison is made among the three initial network classes. Target attractor lengths equal to 500 and 800.}
\label{fig:plot-rld-b}
\end{figure}

We first discuss the results concerning the performance of each class of networks, addressing questions (a), (b) and (c).
The first notable observation is that \textbf{for all} target attractor lengths and \textbf{for all} initial network classes the GA could find at least one network with maximal fitness in the 100 independent runs. This result means that all three classes of networks can be evolved to successfully reach the target. To assess the robustness of the process, we compare the fraction of successful runs at each generation of the algorithm, i.e., we estimate the \textit{success probability at generation t}, defined as the probability that a network with maximal fitness is found at generation $t' \le t$. The corresponding plots are depicted in Figures~\ref{fig:plot-rld-a},~\ref{fig:plot-rld-b}. Results for attractor lengths of 1 and 10 are omitted, because the fraction of runs achieving maximal fitness reaches the 100\% right in the initial population or after few generations.

We first note that the performance achieved with initially ordered networks is considerably lower than that of critical and chaotic ones. This can be ascribed to the fact that ordered networks are not very likely to have long attractors. Anyway, the search process performed by the GA is still able to find a network with the desired attractor length. The case of critical and chaotic networks has some subtleties which deserve to be outlined. First of all, we observe that the success ratio decreases as the attractor length increases. Moreover, in most cases critical networks dominate or are almost equivalent to chaotic ones, while for target attractor length equal to 100, initial chaotic networks seems to provide a better start to the GA. Both the phenomena can be explained by the combination of two factors. First: the cutoff imposed on simulation steps limits from above the networks attractor length, hence making it difficult to evolve networks with an attractor of length comparable with the maximal number of simulation steps because, if an attractor is not found, the corresponding fitness value is zero. Second: critical networks have usually many attractors, but of small length compared to attractor periods of chaotic networks, that can be exponential in the number of nodes. In a survey experimental analysis, we observed that for networks with 100 nodes and a maximal number of simulation steps of 1000, the median attractor length for critical networks is 6, while for chaotic ones is 130. Therefore, for a target length of 100, the fitness of individuals composing the initial population is likely to be higher in the case of chaotic networks than in critical ones. However, it is worth to be noted that critical networks can be anyway evolved to reach long attractors, despite their handicap in the initial population's fitness. This could be a further evidence of their tendency of maximising adaptiveness.
The study of the search space, that would provide insight into problem hardness, is subject of ongoing work.

\subsection{Influence of GA parameters}

\begin{figure}
\begin{center}
 \includegraphics[scale=0.46]{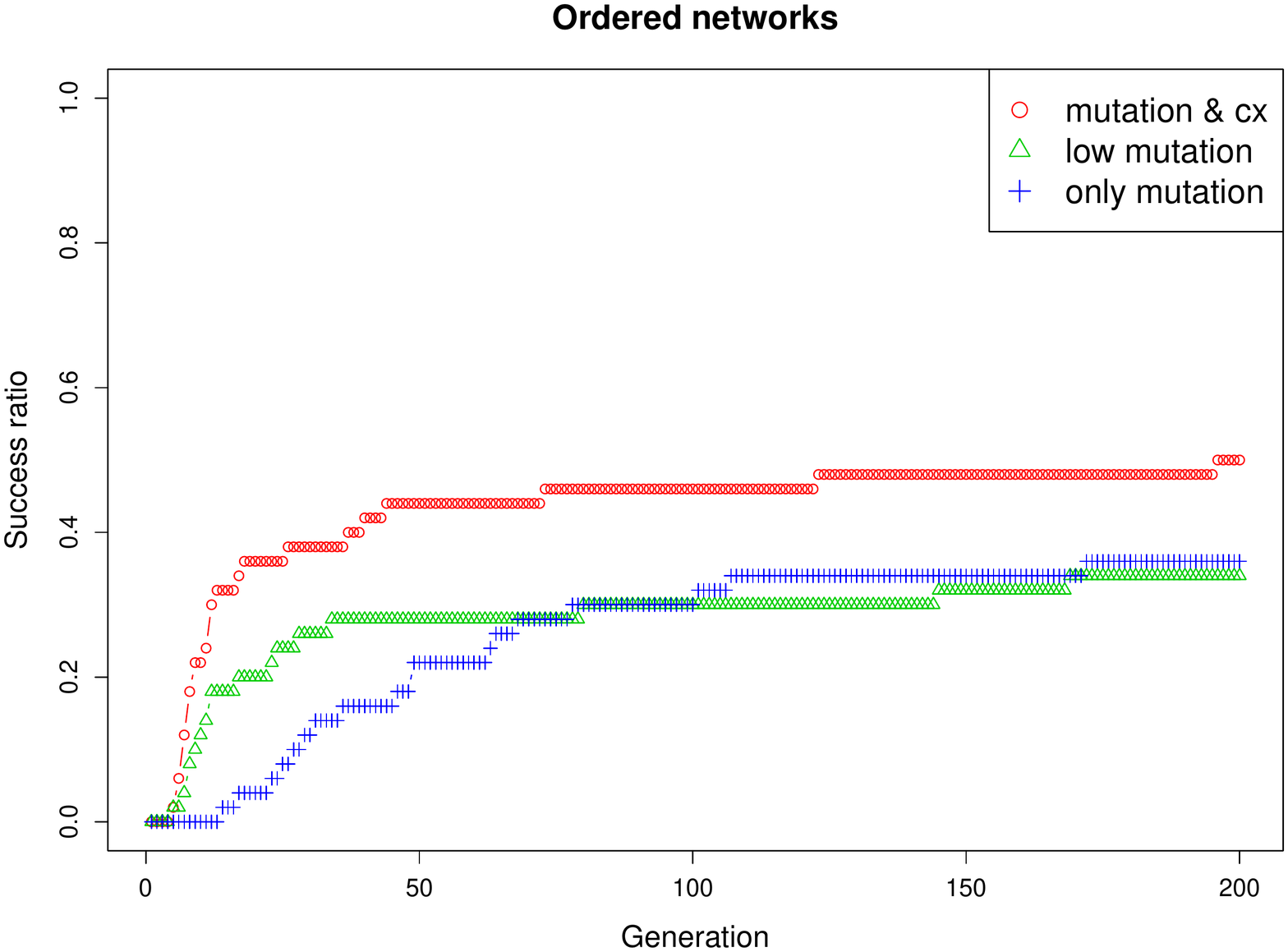}
 \includegraphics[scale=0.46]{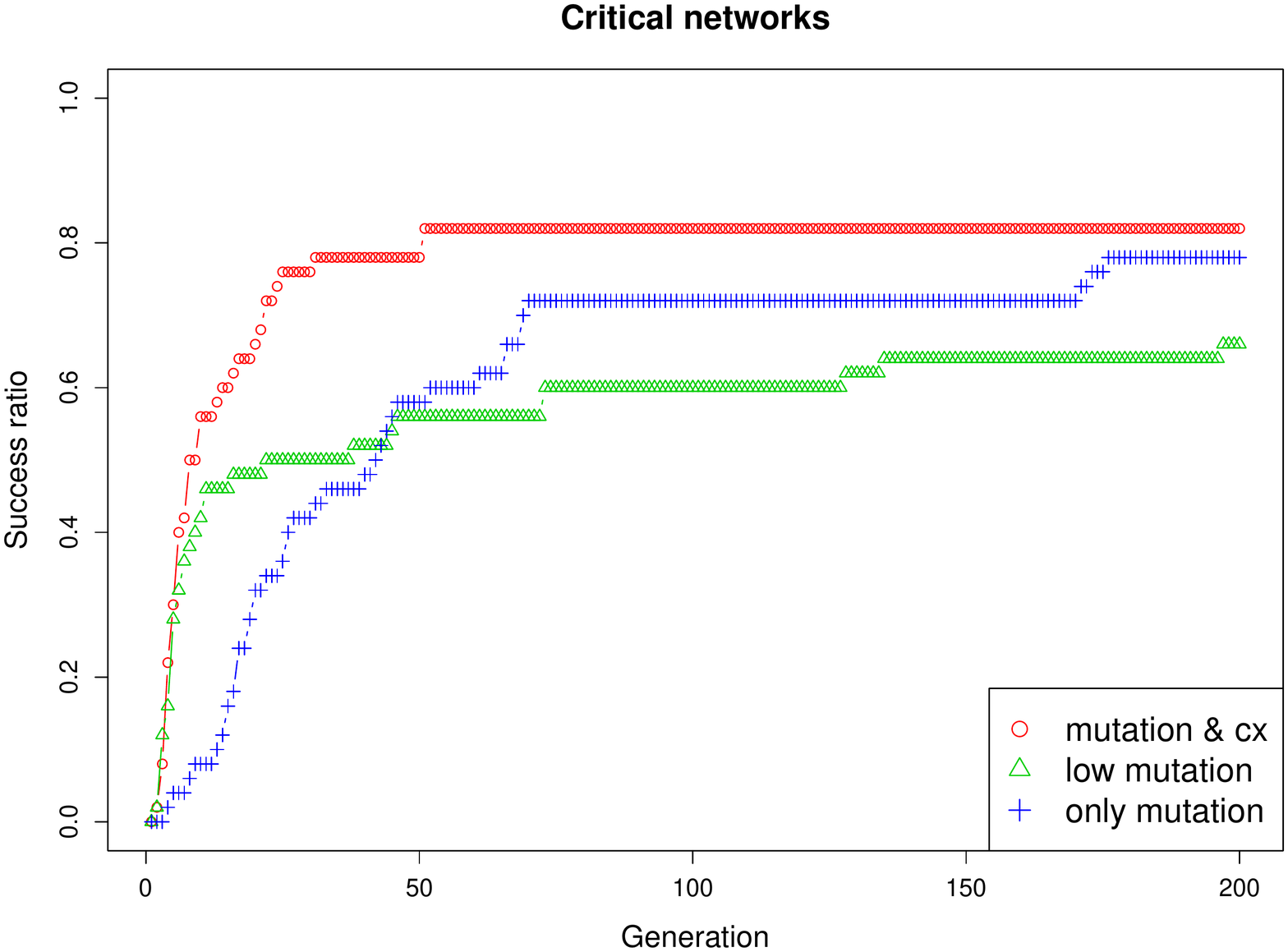}
\end{center}
\caption{Comparison of the impact of mutation and crossover on search performance. The case of ordered and critical initial network classes are reported.}
\label{fig:plot-ga-params-a}
\end{figure}

\begin{figure}
\begin{center}
 \includegraphics[scale=0.46]{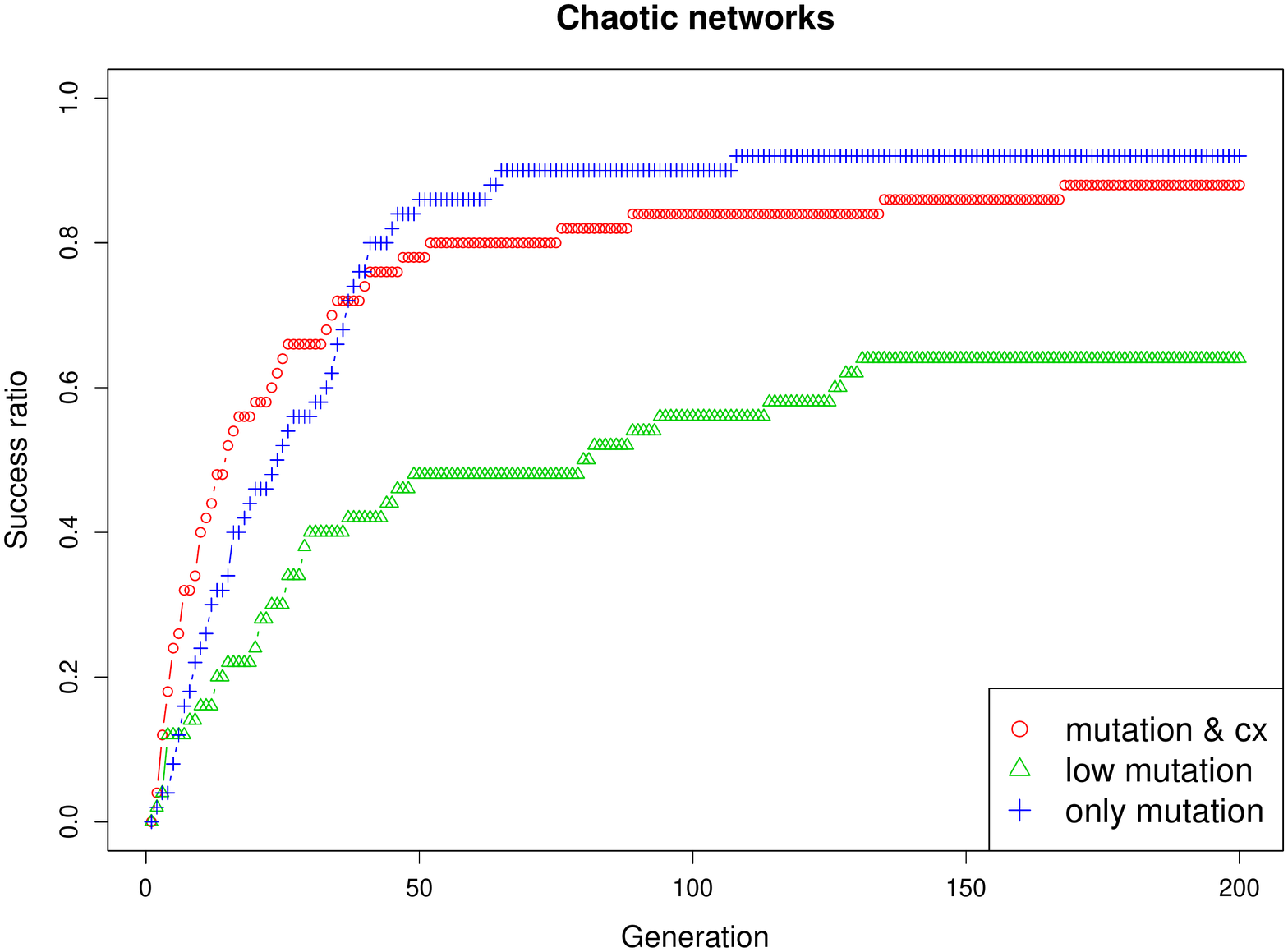}
\end{center}
\caption{Comparison of the impact of mutation and crossover on search performance. The case of initial chaotic network class is reported.}
\label{fig:plot-ga-params-b}
\end{figure}

The influence of mutation and crossover on search performance can shed light on the evolution characteristics of the different initial population classes and can answer question (d).
Figures~\ref{fig:plot-ga-params-a},~\ref{fig:plot-ga-params-b} show a typical case\footnote{Target attractor length equal to 100.} of algorithm performance in the three examined cases of mutation and crossover rates. From the plots we observe that the synergy of both mutation and crossover are crucial for the evolution of initially ordered and critical networks. 
Conversely, for chaotic networks, mutation is much more important than crossover.

\subsection{Effect of evolution on Boolean function homogeneity}

We conclude this analysis by comparing homogeneity distribution at the beginning and at the end of the search. These results should be taken \textit{cum grano salis}, as evolved networks might not have the very same properties as the random initial ones and a complete answer to question (e) requires also to study the properties of Boolean functions as well as network dynamics.
Nevertheless, since only Boolean functions are evolved and topology is kept constant, the evolution of homogeneity can still provide some insights into the effects of evolution on the initial BNs. The average homogeneity of the best individual in the initial and final populations are compared in Table~\ref{tab:p}, where statistically significantly differing averages are denoted by a star.\footnote{i.e., those which passed the Wilcoxon test with confidence level 95\%.}
We can observe a mild tendency of homogeneity decrease for ordered and critical networks, while the GA does not affect homogeneity in chaotic networks. The conclusion we can draw is that, in our experimental setting, the GA does not dramatically change the distribution of 0s and 1s, even if there are some clues suggesting that ordered and critical networks are more affected than chaotic ones and they are somehow pushed towards the chaotic region. However, a more detailed analysis is required before drawing strong conclusions on the effect of GA on network structure.

\begin{table}[t]
\caption{Comparison of the average homogeneity of the best individual in the initial and final population. Significantly differing averages, i.e., those which passed the Wilcoxon test with confidence level 95\%, are denoted by a star.}
\scalebox{0.66}{
\begin{tabular}{|l|c|c||c|c||c|c|}
\hline
Target & \multicolumn{2}{c||}{\textbf{Ordered networks}} & \multicolumn{2}{c||}{\textbf{Critical networks}} & \multicolumn{2}{c|}{\textbf{Chaotic networks}}\\
attr. & \multicolumn{2}{c||}{Best indiv. average homogeneity} & \multicolumn{2}{c||}{Best indiv. average homogeneity} & \multicolumn{2}{c|}{Best indiv. average homogeneity}\\
length & \multicolumn{2}{c||}{initial/final} & \multicolumn{2}{c||}{initial/final} & \multicolumn{2}{c|}{initial/final}\\
\cline{2-7}
 & Successful runs & Unsuccessful runs & Successful runs & Unsuccessful runs & Successful runs & Unsuccessful runs\\
\hline
50 & 0.8474/0.8369* & 0.8447/0.8458 & 0.7807/0.7790 & 0.7895/0.7884  & 0.5023/0.4986 & 0.4943/0.5016\\
100 & 0.8456/0.8345* & 0.8459/0.8380* &  0.7844/0.7752* & 0.7818/0.7822 & 0.5032/0.5034  & 0.4964/0.4996\\
500 & 0.8434/0.8285  & 0.8465/0.8365* & 0.7860/0.7688* & 0.7824/0.7735* & 0.5019/0.5046  & 0.5010/0.5018\\
800 & 0.8325/0.7962  & 0.8463/0.8346* &  0.7854/0.7770 & 0.7834/0.7700* & 0.4951/0.5034  & 0.4997/0.5039\\
\hline
\end{tabular}
}
\label{tab:p}
\end{table}

\section{Conclusion and outlook to future work}
\label{sec:agenda}

In this work, we have presented and discussed results of the evolutionary design of BNs with a desired attractor length. We have shown that it is possible to find networks with such a property for every kind of initial class: ordered, chaotic and critical. Search performance starting from critical and chaotic networks is considerably higher that in the case of ordered networks. Another important outcome of the experiments is that, critical networks are a good start for GA for all the target attractor lengths tested, despite the low probability of finding long attractors in those networks.
This work is just a first step in this research area. 
A future research agenda include: \textit{(i)} relaxing the constraint of keeping constant the initial state, thus moving to stochastic search problems (as the enumeration of all possible initial states is impractical), \textit{(ii)} evolving also network topology and exploring the use of different search algorithms, mainly metaheuristics and their hybrids. Finally, we are also planning to experiment with other targets, such as specific patterns in the attractors and combinations thereof, aiming at the design of networks with a desired landscape of attractors, each with a specific characteristic.

\section*{Acknowledgements}

We thank Roberto Serra, Marco Villani for useful comments and suggestions.


\end{document}